\pdfoutput=1
\documentclass[11pt]{article}

\usepackage[final]{acl}

\usepackage{times}
\usepackage{latexsym}

\usepackage[T1]{fontenc}
\usepackage[utf8]{inputenc}

\usepackage{microtype}

\usepackage{inconsolata}

\usepackage{caption}
\usepackage{subcaption}
\usepackage{pgfplots}
\usepackage{booktabs} 
\usepackage{tikz}
\usepackage{bbding} 
\usepackage{xcolor}
\usepackage{adjustbox}
\usetikzlibrary{positioning, fit, arrows.meta, shapes.geometric}
\usepackage{amsmath}
\usepackage{amsfonts}

\DeclareMathOperator*{\argmin}{arg\,min}
\DeclareMathOperator*{\E}{\mathbb{E}}

\title{DIEK\AE: Difference Injection for Efficient Knowledge Augmentation and Editing of Large Language Models}

\author{Alessio Galatolo, Meriem Beloucif$^*$, Katie Winkle\thanks{Equal supervision.} \\
  Uppsala University\\
  \texttt{\{alessio.galatolo, katie.winkle\}@it.uu.se}; \\ \texttt{meriem.beloucif@lingfil.uu.se} \\}

\begin{document}
\maketitle
\begin{abstract}
Pretrained Language Models (PLMs) store extensive knowledge within their weights, enabling them to recall vast amount of information. However, relying on this parametric knowledge brings some limitations such as outdated information or gaps in the training data. This work addresses these problems by distinguish between two separate solutions: knowledge editing and knowledge augmentation. We introduce Difference Injection for Efficient Knowledge Augmentation and Editing (DIEK\AE), a new method that decouples knowledge processing from the PLM (LLaMA2-7B, in particular) by adopting a series of encoders. These encoders handle external knowledge and inject it into the PLM layers, significantly reducing computational costs and improving performance of the PLM. We propose a novel training technique for these encoders that does not require back-propagation through the PLM, thus greatly reducing the memory and time required to train them. Our findings demonstrate how our method is faster and more efficient compared to multiple baselines in knowledge augmentation and editing during both training and inference. We have released our code and data at \url{https://github.com/alessioGalatolo/DIEKAE}.
\end{abstract}

\section{Introduction}
Pretrained Language Models (PLMs) possess the capacity to encode and preserve the knowledge from their training within their weights \cite{carlini2021extracting}. Previous works have even introduced the concept of knowledge neurons \cite{dai-etal-2022-knowledge}, pointing to the idea that knowledge is stored in the Multi-Layer Perceptron (MLP) part of the Transformer architecture \cite{vaswani2017attention}, with \cite{dai-etal-2022-knowledge} and multiple subsequent works showing how to edit specific facts in these layers \cite{dai-etal-2022-knowledge, geva-etal-2021-transformer, meng2022rome, meng2023memit, li2024pmet}.

However, relying only on this parametric knowledge has some flaws: knowledge can become outdated (e.g. `The current president of the United States') and is limited by what was included in the training data, where very specialised or specific knowledge may have been excluded. These flaws have brought two emerging tasks and research directions in Natural Language Processing (NLP): `knowledge augmentation' and `knowledge editing'. These approaches are especially relevant for smaller-sized PLMs, where embedded knowledge is severely constrained by size and parameter count \cite{roberts-etal-2020-much, kaplan2020scaling, allen2024physics}.

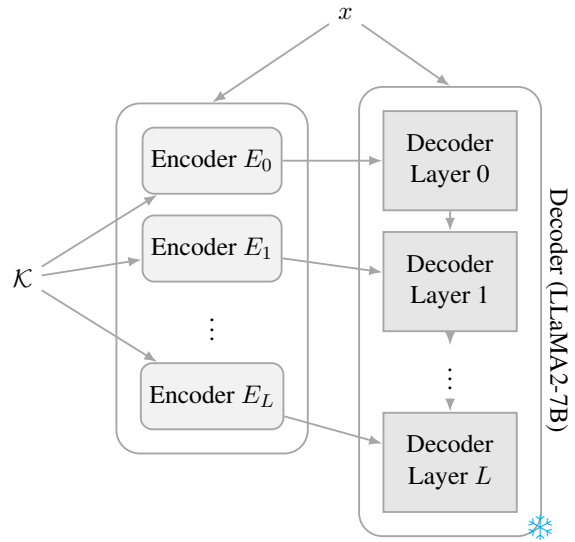
\begin{figure}
    \centering
    \adjustbox{max width = \columnwidth}{
    \begin{tikzpicture}[
        node distance=1.0cm and 1.5cm,
        box/.style={draw=gray!70, fill=gray!10, thick, minimum width=2cm, minimum height=1cm, align=center, rounded corners=5pt},
        decoder/.style={draw=gray!70, fill=gray!20, thick, minimum width=2cm, minimum height=1cm, align=center},
        arr/.style={-{Latex[length=2mm]}, thick, gray!70},
        dot/.style={circle, inner sep=0pt, minimum size=2pt, fill=gray!70},
        ]

        % Knowledge
        \node (knowledge) {$\mathcal{K}$};

        % Encoders
        \node[box, above right=of knowledge] (encoder0) {Encoder $E_0$};
        \node[box, below=0.3cm of encoder0] (encoder1) {Encoder $E_1$};
        \node[below=0.2cm of encoder1] (ellipsis) {$\vdots$};
        \node[box, below=0.2cm of ellipsis] (encoder3) {Encoder $E_L$};
        
        % Decoder
        \node[decoder, right=of encoder0, minimum height=1.5cm] (decoder0) {Decoder\\Layer 0};
        \node[decoder, below=0.3cm of decoder0, minimum height=1.5cm] (decoder1) {Decoder\\Layer 1};
        \node[below=0.2cm of decoder1] (ellipsis2) {$\vdots$};
        \node[decoder, below=0.2cm of ellipsis2, minimum height=1.5cm] (decoder3) {Decoder\\Layer $L$};
        
        % Draw big rectangle around decoder blocks
        \node[draw=gray!70, thick, fit=(decoder0) (decoder3), inner sep=10pt, rounded corners=10pt, label={[label distance=-2cm, text depth=3ex, rotate=-90]right:Decoder (LLaMA2-7B)}] (decoderbox) {};
        
        % Draw big rectangle around encoder blocks
        \node[draw=gray!70, thick, fit=(encoder0) (encoder3), inner sep=10pt, rounded corners=10pt, label={[label distance=-2cm, text depth=3ex, rotate=-90]right:}] (encoderbox) {};
        
        % Additional nodes and arrows
        \node[above=1.5cm of encoder0, xshift=2cm] (x) {$x$}; % New node for "x"
        \draw[arr] (x) -- (decoderbox.north); % Arrow to decoderbox
        \draw[arr] (x) -- (encoderbox.north); % Arrow to encoder0

        % Connections
        \foreach \i in {0,1,3}
            \draw[arr] (knowledge) -- (encoder\i);
        \foreach \i in {0,1,3}
            \draw[arr] (encoder\i) -- (decoder\i);

        \draw[arr] (decoder0) -- (decoder1);
        \draw[arr] (decoder1) -- (ellipsis2);
        \draw[arr] (ellipsis2) -- (decoder3);

        % Add snowflake symbol to bottom right of decoderbox
        \node[anchor=south west, xshift=-10pt, yshift=-5pt] at (decoderbox.south east) {\textcolor{cyan}{\SnowflakeChevron}};
    \end{tikzpicture}
    }
    \caption{Illustration of our method. $x$ is the input and $\mathcal{K}$ is some external knowledge. \textcolor{cyan}{\SnowflakeChevron}\; indicates frozen PLM.}
    \label{fig:method}
\end{figure}

In this work, we distinguish the two tasks using the following definitions. In both of them, the goal is to update the language model's generations to reflect a new set of knowledge $\mathcal{K}$. And, while the former aims at `augmenting' the generations with specific knowledge, conditioning them at inference time; the latter focuses on editing the intrinsic knowledge at an earlier time, often by updating a subset of the model's weights, in a way that can be used \textit{directly} at inference time (i.e. without further providing the knowledge). 

More specifically, they can be distinguished based on their inference objective. Consider a decoder-only\footnote{While the objective can be generalised to all language models, in this work we focus on decoder-only auto-regressive models, so we drop the generic notation $f$ to indicate the model and instead use $D$.} language model $D$ with pretrained weights $\theta$ and a distribution of knowledge problems $(x_i, y_i, \mathcal{K}_i) \sim p(\mathcal{D})$ where $x_i$ is a given input, $\mathcal{K}_i$ is some knowledge and $y_i$ is the desired output that follows from that knowledge input. 

A knowledge or model editing objective is like the following:
\begin{equation}
    \argmin_{\hat{\theta}} \E_{x_i, y_i, \mathcal{K}_i \sim p(\mathcal{D})} [\mathcal{L}(D_{\hat\theta}(x_i), y_i)]
\end{equation}
Where $\hat\theta$ are new weights (or edited weights) to reflect or memorise the knowledge $\mathcal{K}$.

On the other hand, we can formulate a knowledge augmentation objective as the following:
\begin{equation}
    \argmin_{\hat{\theta}} \E_{x_i, y_i, \mathcal{K}_i \sim p(\mathcal{D})} [\mathcal{L}(D_{\hat\theta}(\mathcal{K}_i, x_i), y_i)]
\end{equation}
Where $\hat\theta$ are new weights optimised to be able to make use of a generic knowledge $\mathcal{K}$ (when presented at inference time).

Note also that while the knowledge $\mathcal{K}$ in knowledge editing works is often restricted to simple sentences of the type <subject, relation, object> \cite{wang2023knowledge}, the knowledge $\mathcal{K}$ used for knowledge augmentation is often very long and can contain many (even irrelevant) facts. 

Works in knowledge augmentation approach the problem by including the knowledge in the context window of the model, and letting the model discern the correct output \cite{pmlr-v119-guu20a,clark2021transformers, dinan2018wizard, lewis2020retrieval}. Following \citet{chen2023reckoning}, we refer to this ability as in-context reasoning (ICR). These works then can focus on the retrieval of the knowledge to be included, selecting a number of sentences (or chunks) that can fit in the context window, and take the name of Retrieval Augmented Generation (RAG) methods \cite{lewis2020retrieval}. The main downsides of these approaches are 1) context windows are often limited to a few tokens (e.g. 2048 for LLaMA, \citet{touvron2023llama}; and 4096 for LLaMA2, \citet{touvron2023llama2}) and 2) as also noted by \citet{hu2024survey}, the associated computational costs, which scale quadratically with the size of the context window \cite{vaswani2017attention}.

In this work, we first aim to solve the knowledge augmentation problem by complementing a decoder-only PLM $D$ with a series of small encoders $E=\{E_0, ..., E_L\}$ (Figure \ref{fig:method}). The encoders take the knowledge $\mathcal{K}_i$ in place of the decoder alongside the input $x_i$. Their output $E(\mathcal{K}_i, x_i)$ is injected into the different layers $D_0, ..., D_L$ of the decoder $D$, where $L+1$ is its number of layers/Transformer blocks. Thanks to its restricted size, having the encoder process the knowledge is much less costly and can yield very similar, if not better, results than providing it to the decoder. For the decoder-only PLM we choose LLaMA2-7B \cite{touvron2023llama2} and we keep its weights frozen for all of our experiments.

We then develop and test a new training objective for the initial training of the encoders $E$ that does not require gradient computations of $D$ but only relies on its hidden states collected through two forward passes.

Finally, we also show how our method can be adapted to the other task of knowledge editing. Given that our encoder is trained to propagate its knowledge to the PLM, we can edit its weights at much less of the cost of editing the PLM's and the changes will be propagated to the PLM at inference time \textit{without} the need of providing the knowledge again: $E(\_, x_i)$.

We can summarise our contributions as the following:
\begin{enumerate}
    \item We introduce Difference Injection for Efficient Knowledge Augmentation and Editing (DIEK\AE), a method where we decouple knowledge from the PLM, delegating its processing and injection to separate encoders.
    \item We show how our method brings substantial computational benefits both during training and inference.
    \item We propose a novel training objective for knowledge augmentation that does not require back-propagating through the PLM but only through our encoders, thus greatly reducing training requirements.
\end{enumerate}
\section{Related Works}
There are a number of related methods that are relevant to this work both in knowledge editing and knowledge augmentation. But, maybe more interestingly, some works fall in the grey area between the two. Following \citet{wang2023knowledge, wei2024stable} we refer to these methods as memory-based.
\subsection{Knowledge Augmentation}
Most works in knowledge augmentation directly add the knowledge to the decoder's original input or to its hidden states \cite{huang2024survey, hu2024survey}, circumventing entirely the idea of enhancing the architecture or adding knowledge encoders. \citet{lewis2020retrieval} develop a fine-tuning technique to augment the generations of a PLM at inference time by including both parametric knowledge and snippets retrieved from Wikipedia, effectively giving birth to a new strand of research on RAG methods.

There are few works that explore the idea of separate encoders for knowledge \cite{sridhar-yang-2022-explaining, wang-etal-2021-k}. \citet{wang-etal-2021-k} train a number of adapters\footnote{Whilst using the terminology of `adapters', the authors are effectively using a derivative of the Transformer architecture \cite{vaswani2017attention}, which thus can be considered similar to what we call encoders in this work.} to augment a given PLM, each adapter is aimed at different kinds of knowledge (e.g. Wikipedia, linguistic knowledge, etc.) and can be used on its own or with other adapters. At inference time, the output of the adapter(s) is concatenated to (some) hidden states of the PLM. \citet{sridhar-yang-2022-explaining} train different expert knowledge models whose output is joint with the addition of separator tokens. Both these works concatenate processed knowledge to create the input for the decoder, and, while this brings improved performance, it also adds a considerable computational overhead. 

\citet{dai2023neural} explore the idea of a `neural knowledge bank', where additional knowledge is store and then injected, still by concatenation, in the MLP layers. 

What differentiates our work from these previous contributions is that our encoder-processed knowledge is directly injected into various layers of the decoder simulating a knowledge-aware run without actually increasing the length/size of the input/hidden states, thus greatly improving computational costs without sacrificing the PLM performance.
\subsection{Knowledge Editing}
ROME \cite{meng2022rome} and MEMIT \cite{meng2023memit} are two works that base their method on identifying and editing the layers responsible for a given fact. Following their localisation they edit the 5th MLP layer (ROME) and MLP layers $\{3, 4, 5, 6, 7, 8\}$ (MEMIT). MEMIT represents an evolution of ROME that is able to adapt to several thousands of edits (authors show up to 10k) compared to the few of ROME.

\citet{li2024pmet} expand on these ideas by also analysing the multi-head self-attention achieving a more precise editing. \citet{yu2024melo} show how LoRA \cite{hu2022lora} can be used for effective model editing. \citet{wei2024stable} develop a new data augmentation technique for training with a knowledge editing objective, where paraphrases and different contexts are included in the prompt.

Multiple works \cite{meng2023memit, yu2024melo, li2024pmet, yao-etal-2023-editing} have also shown that simple fine-tuning (FT) does not generally perform well for the task of knowledge editing performs worse and is generally used as a weak baseline.
\subsection{Memory-Based Knowledge Editing}
\citet{pmlr-v162-mitchell22a} introduce SERAC, a method to augment a given PLM $f$ with a newly trained `counterfactual model'. At inference time, if the input refers to some knowledge that was updated, the counterfactual model would reply in place of the main one. This work, of course, has the downside of having to train a completely separate model whose performance is often suboptimal compared to the main one's.

Another work in the intersection of these two approaches is that of \citet{chen2023reckoning}. Here, the authors target logical reasoning where the knowledge is represented as a collection of facts and rules and the input is a statement that can be `True', `False' or `Unknown' based on the knowledge. They develop a method that learns new weights for a given model that also encodes the given rules and facts. While this work should fall under the knowledge editing umbrella, as the knowledge is not presented during inference, its memorisation in the weights happens on a per-sample basis i.e. they learn optimal meta-parameters that are then adapted to each sample separately.

\section{Methodology}
For our method (shown in Figure \ref{fig:method}),
we take inspiration from Adapters \cite{lin-etal-2020-exploring, wang-etal-2021-k} where additional parameters/embeddings are added to specific layers of the Transformer for adaptation to different downstream tasks. In particular, we use a \textit{series} of encoders $E=\{E_0, ..., E_L\}$, one per Transformer layer in the decoder. In our method, the encoders are the only ones to receive the knowledge, keeping the decoder's context short and focused to the input. In doing so, we keep the decoder's weights frozen and train the encoders to inject the given knowledge into the different layers of the decoder.

Each encoder's output $E_l(\mathcal{K}, x)$, where $l \in \{0,...,L\}$, is added to the that of the corresponding layer $D_l$ of the Transformer: 
\begin{equation}
    \bar{D}_l(x) = D_l(x) + shift_\text{left}(E_l(\mathcal{K}, x), |\mathcal{K}|))
\end{equation}
$shift_\text{left} (a, b)$ is an operator that shifts $a$ by $b$ positions. Because in the encoder(s) we prepend the knowledge to the input, the beginning of the input in $D$ is at position $0$, while in $E$ is at position $|\mathcal{K}|$, by shifting the output of $E$ towards the left, we ensure that the two positions match. The initial $|\mathcal{K}|$ values get discarded and are \textit{not} added to the decoder's hidden state. When including the knowledge in the prompt, we use the template from \citet{cohen2024ice} which surrounds the knowledge with ``Imagine that \{ <knowledge> \}''.

In general, our approach is intended to be used with one encoder per layer, however, we also experiment with a lower number of encoders (therefore injecting fewer layers of $D$), resulting in improved performance \textit{and} computational costs (see Section \ref{sec:less_encoders}).

\subsection{Encoder}
\label{sec:encoder}
\begin{figure}
    \centering
    \begin{tikzpicture}[
        arr/.style={-{Latex[length=2mm]}, thick, gray!70},
        box/.style={draw=gray!70, fill=gray!10, thick, minimum width=2cm, minimum height=1cm, align=center, rounded corners=5pt}
    ]
    % Rectangle with rounded corners
    \node [box, align=center] (encoder) at (0,0) {Encoder $E_l$};
    
    % Down-projection layer (funnel-shaped trapezoid)
    \node [trapezium, draw=gray!70, fill=blue!20, thick, trapezium left angle=110, trapezium right angle=110, minimum width=3cm, minimum height=1cm, align=center] (downproj) at (0,1.5) {Down-projection\\ layer};
    \draw [arr] (downproj) -- (encoder);
    
    % Embeddings(x) text
    \node [above] at (0,2.5) {Embeddings($\mathcal{K}, x$)};
    \draw [arr] (0,2.5) -- (downproj);
    
    % Up-projection layer (funnel-shaped trapezoid)
    \node [trapezium, draw=gray!70, fill=blue!20, thick, trapezium left angle=70, trapezium right angle=70, minimum width=3cm, minimum height=1cm, align=center] (upproj) at (0,-1.5) {Up-projection \\layer};
    \draw [arr] (encoder) -- (upproj);
    
    \end{tikzpicture}
    \caption{Illustration of a single encoder.}
    \label{fig:encoder}
\end{figure}
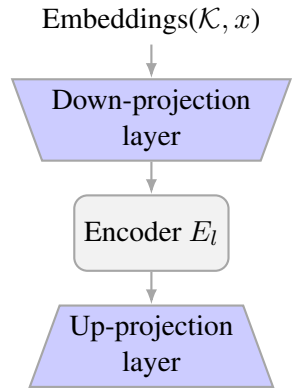

For the encoder, we choose the same architecture as our PLM of choice, LLaMA2 \cite{touvron2023llama2}. Each encoder is made up of four Transformer blocks of 128 dimensions with no other meaningful change. Given the reduced hidden dimension, we also introduce two projection layers that are placed at the beginning and the end of the encoder: one takes the token embeddings that we obtain using LLaMA's ones and `down'-projects them to be used by the encoder; the other one `up'-projects the output of the encoder to be used in the decoder. We illustrate this in Figure \ref{fig:encoder}. Each encoder contains about 2M parameters, $<0.03\%$ of the original PLM.

Note how we do not experiment with architectural or hyperparameter changes in the encoder. Our goal is to show that, with our new method and training procedure, the encoder's performance would scale in the same way as larger models do, while saving the training time.

\subsection{Training}
\label{sec:training}
Fine-tuning large language models is an operation that is expensive both in time and space. As \citet{malladi2023finetuning} report, fine-tuning can take as much as $12\times$ the GPU space needed for inference, requiring as much as 80GB of GPU memory for 2.7B models. For this reason, multiple techniques have been developed for Parameter-Efficient Fine-Tuning (PEFT) such as LoRA \cite{hu2022lora} or QLoRA \cite{dettmers2023qlora}. In our case, given that we want to condition the outputs of the PLM, we would have to build the gradient graph for the whole model, requiring almost the same resources as full fine-tuning. For this reason, we develop and test a new training objective that does not rely on back-propagating through the decoder $D$, but only computes the gradients for each encoder $E_l$ where its objective is extracted from two forward passes of the decoder.

Our training objective comes from the intuition that ICR (in-context reasoning) yields better results than not including any knowledge in the prompt i.e. $p(y|\mathcal{K}, x) > p(y|x)$ given input $x$, target $y$ and knowledge $\mathcal{K}$. With this idea in mind, we want our encoders $E$ to approximate the run with the knowledge, without actually feeding it to the decoder. With a slight abuse of notation, we define our goal as $p(y|E(\mathcal{K}, x) + x) \approx p(y|\mathcal{K}, x)$.

To achieve this, we run two forward passes of $D$ and save the hidden states $o_l$ of each intermediate layer $l \in \{0, 1, ..., 31\}$ (32 total layers for LLaMA2-7B), one pass is done with only the input $x$ while in the second we prepend the knowledge $\mathcal{K}$ to the input. More specifically we collect $o_i^{(x, \mathcal{K})}$ from $D(\mathcal{K}, x)$ and $o_l^{(x)}$ from $D(x)$. Then, each encoder $E_l$ relative to layer $l$ is trained to predict the difference in the hidden states using a mean squared error (MSE) loss:

\begin{equation*}
    \mathcal{L}_\text{MSE} [E_l(\mathcal{K}, x), o_l^{(x, \mathcal{K})} - shift_\text{right}(o_l^{(x)}, |\mathcal{K}|)]
\end{equation*}
Here, $shift_\text{right}$ acts exactly like $shift_\text{left}$ except that it shifts to the right (missing values are not computed in the loss).

Note that whilst the idea of training or fine-tuning with just forward passes is not entirely novel \cite{malladi2023finetuning}, our work is very different from that of e.g. Zeroth-Order optimisation \cite{spall1992multivariate} where a model is optimised by doing multiple forward passes with the same input but adding perturbations to the weights of the model. In our case, we still use only forward passes, preserving the memory gains but we do so by varying the input (and keeping the weights fixed). Further, \citet{malladi2023finetuning} prove that these class of methods work only for fine-tuning\footnote{The authors show that the method converges in a reasonable amount of time under the assumption that the language model has been extensively trained and that the fine-tuning objective is similar to that of pretraining.}, while ours is specifically targeted at pretraining.

\subsection{Fine-Tuning}
We couple (and compare) the above training objective with further fine-tuning (of the encoders only) for specific datasets/tasks, using the following objective:
\begin{equation*}
    \mathcal{L}_\text{CE} [D(E(\mathcal{K}, x) + x), y]
\end{equation*}
Where $\mathcal{L}_\text{CE}$ is the Cross-Entropy loss and, for simplicity, we write $D(E(\mathcal{K}, x) + x)$ as meaning to add the output of each $E_l$ to the hidden state of the corresponding layer. 

\subsection{Knowledge Editing}
While the main focus of this work is knowledge augmentation, we also show how our method can be easily adapted to the problem of knowledge editing. This is based on the assumption that editing the weights of our encoders is much less costly than editing those of the PLM itself and that the encoders would be able to easily propagate the knowledge to the PLM. For this, we use the objective defined in Equation 1, which, adapted to our method becomes:
\begin{equation*}
    \mathcal{L}_\text{CE} [D(E(\_, x) + x), y]
\end{equation*}
Note how, in this case, neither the decoder nor the encoders receive any knowledge, instead relying on changes in the weights to optimise the objective.

\section{Experiments}
Our experiments all begin with a common training, following the objective described in Section \ref{sec:training}. We then evaluate the performance of our method on the tasks of knowledge augmentation and knowledge editing using three different mixtures of datasets. We further experiment with fine-tuning our encoders to each one of the datasets. We keep the decoder's weights frozen \textit{at all times}.

We provide the code for our method, dataset preprocessing and various experiments at \url{https://github.com/alessioGalatolo/DIEKAE}.
\subsection{Knowledge Augmentation Datasets}
We use six datasets for `general' training. CMU Dog \cite{zhou-etal-2018-dataset} contains conversations between two users (source/target) based on a given snippet of film information (knowledge). Curiosity \cite{rodriguez-etal-2020-information} contains conversations between an assistant (target) and a user (source), the assistant replies to the user's queries by including relevant `fun' facts (knowledge). DREAM \cite{sun-etal-2019-dream} is a reading comprehension dataset whose text, containing conversations between two people (knowledge), is paired with questions (source) and answers (target). Natural Questions \cite{kwiatkowski-etal-2019-natural} contains open-domain knowledge-based question-answering (source-target). QUASAR-T \cite{dhingra2017quasar} is another question-answering (source-target) dataset whose answers are based on retrieved web searches (knowledge). Wizard of Wikipedia \cite{dinan2018wizard}, as the name suggests, contains conversations between a user (source) and a `wizard' (target) who bases their answers on Wikipedia's pages content. 

Among these six datasets used for training, we exclude Natural Questions \cite{kwiatkowski-etal-2019-natural} during evaluation as the version we adopt does not contain a test set.
\subsection{Datasets for ICR and Knowledge Editing}
For the task of ICR and Knowledge Editing we adopt three new datasets that have been previously used for knowledge editing works. ProofWriter \cite{tafjord-etal-2021-proofwriter} and FOLIO \cite{han2022folio} were used by \citet{chen2023reckoning} to evaluate reasoning from rules and facts. Here, the authors use a method to memorise the rules and facts in the model's parameters and then use the newly learned parametric knowledge to classify statements into True/False/Unknown. The process is done on a per-sample basis, where further tuning is needed for each new set of knowledge. 

CounterFact, created and used by \citet{meng2023memit} is a collection of false facts that are used to edit the parametric knowledge of PLMs. The authors take into consideration more datasets than this one, but we restrict our attention to it as it is the only one that does not overlap with knowledge already present in the PLM. As the name suggests, the dataset contains facts that are effectively false e.g. ``The capital of Belgium is Paris''.

\subsection{Baselines}
For each of our experiments, we adopt different baselines: 
\begin{enumerate}
    \item Plain input fed to the PLM, $D(x)$.
    \item Knowledge and input fed to the PLM, $D(\mathcal{K}, x)$.
    \item Fine-tuned PLM using LoRA \cite{hu2022lora} with the objective of making use of the knowledge given at inference time.
    \item Fine-tuned PLM through traditional supervised fine-tuning (SFT) with the objective of making use of the knowledge given at inference time.
\end{enumerate}
For the first two baselines, we do not do any training, while both LoRA and SFT are trained using the same data as our method. All the methods (including our own) are used with a context window of 1024 tokens, which is enough to contain (almost) all of the inputs as well as the knowledge. For the implementation of the baselines as well as our method we rely on the transformers library \cite{wolf-etal-2020-transformers}.

\section{Results}
Our results show that our fine-tuning method yields the best results when only a subset of the encoders/layers are used. For this reason, when in this section we refer to our method (fine-tuned), we specifically only used and fine-tuned a few of the encoders. An analysis of the effects of different subsets of encoders is presented in Section \ref{sec:less_encoders}. Surprisingly, we find that the best results are achieved when using the same subset of layers identified by \citet{meng2023memit}: $\{3, 4, 5, 6, 7, 8\}$.

\subsection{Knowledge Augmentation}

\begin{table*}[h!]
    \centering
    \adjustbox{max width=\textwidth}{
        \begin{tabular}{@{} lccccccccc @{}}
            \toprule
            & \multicolumn{6}{c}{\textbf{Perplexity $\downarrow$}} & \multicolumn{2}{c}{\textbf{Time (GPU-hours$^\dagger$) $\downarrow$}} & \textbf{Time$^{\dagger\dagger}$ (s) $\downarrow$}\\
            \cmidrule(lr){2-10}
            \textbf{Method} & CMU dog & Curiosity & DREAM & QUASAR-T & WoW & \textbf{All} & Training, total & \textbf{Per epoch} & Inference\\
            \midrule
            LLaMA2: $D(x)$ & 1.20 $\times 10^6$ & 2.42 $\times 10^6$ & 6.25 $\times 10^4$ & 6.00 $\times 10^5$ & 7.89 $\times 10^5$ & 6.13 $\times 10^5$ & - & - & \textbf{238.56}\\
            LLaMA2: $D(\mathcal{K}, x)$ & 5.56 & 3.10 & 6.61 & 9.90 & 4.39 & 5.48 & - & - & 416.15\\
            LoRA: $D_{lora}(\mathcal{K}, x)$ & 4.13 & 2.07 & \textbf{1.78} & \textbf{1.68} & \textbf{2.85} & 2.36 & 52.32 & 22.55 & 423.67\\
            SFT: $D_{sft}(\mathcal{K}, x)$ & \textbf{2.63} & \textbf{2.02} & 1.88 & 1.80 & 3.00 & \textbf{2.22} & 68.99 & 36.31 & 413.14\\
            \midrule
            Ours (32 encoders) & 33.23 & 18.80 & 12.46 & 22.81 & 18.98 & 20.21 & 59.18 & 59.18 & 668.23\\
            Ours (6 encoders) & 5.86 & 3.55 & 7.01 & 11.34 & 4.90 & 6.05 & 14.72 & 14.72 & 268.53\\
            Ours (6 encoders, fine-tuned) & \textbf{\textit{5.11}} & \textbf{\textit{2.89}} & \textbf{\textit{3.20}} & \textbf{\textit{2.18}} & \textbf{\textit{3.99}} & \textbf{\textit{3.33}} & \textbf{11.17} & \textbf{13.3} & \textbf{\textit{268.50}}\\
            \bottomrule
        \end{tabular}
    }
    \caption{Results of our method and comparison with baselines. For each dataset, we take 1000 samples.\footnotesize{$^\dagger$}Normalised against a NVIDIA A40 GPU. $^{\dagger\dagger}$Tested on a local NVIDIA RTX 3090.}
    \label{tab:results}
\end{table*}
We show in Table \ref{tab:results} the average perplexity of our method and baselines in the augmentation task, where knowledge is supplied at inference time. We can see that our hypothesis of knowledge-aware generation yielding better results (than not including knowledge) holds true when considering all the datasets. Further, our training procedure proves very effective as it closely approximates the performance of the PLM with the knowledge provided, although only surpassing it when further fine-tuned. Interestingly, the pretrained-only method performs better when only a selection of the encoders is used, we attribute this result to some of the encoders that were not able to converge during training, having a final loss of above $10^5$ (while most encoders finished the training with a loss of less than ten). Specifically, the diverging encoders are relative to the decoder's layers $\{1, 29, 30, 31\}$, possibly indicating a high variance in hidden states values.

Our results show that our method is competitive with the PLM's knowledge-aware output, while still requiring about half inference time. However, our method falls short when compared to LoRA or SFT. While this might be seen as a negative result, it must be highlighted how these two baselines both leverage the full capabilities of the PLM and as such cannot and \textit{should not} be outperformed by much smaller encoders such as the ones we use. Further, it is to note how our method only needs $1/5$ of LoRA's training time or $1/7$ of SFT's.

\subsection{In-Context Reasoning}
We show in Table \ref{tab:results_proofwriter} the results on ProofWriter and FOLIO. We see that our method is able to improve over the plain PLM, however, it again struggles when compared to the other baselines. Compared to the knowledge augmentation task, the gap is much wider, possibly indicating that, while smaller encoders are `good enough' to inject general knowledge, their restricted size is unable to complete complex reasoning on its context.

\begin{table}[h!]
    \centering
    \adjustbox{width=\columnwidth}{
        \begin{tabular}{@{} lcc @{}}
            \toprule
            & \multicolumn{2}{c}{\textbf{Acc} $\uparrow$} \\
            \cmidrule(lr){2-3}
            \textbf{Method} & Proofwriter & Folio\\
            \midrule
            LLaMA2 - $D(x)$ & 45.2 & 35.3\\
            LLaMA2 - $D(\mathcal{K}, x)$ & 45.4 & 38.73\\
            LoRA - $D_{lora}(\mathcal{K}, x)$ & 98.1 & \textbf{71.57}\\
            SFT - $D_{sft}(\mathcal{K}, x)$ & \textbf{98.9} & 69.12\\
            \midrule
            Ours (fine-tuned) & 56.5 & 41.14\\
            \bottomrule
        \end{tabular}
    }
    \caption{Classification accuracy of our methods on ProofWriter and FOLIO, with comparison on baselines.}
    \label{tab:results_proofwriter}
\end{table}

\subsection{Knowledge Editing}
To evaluate knowledge editing, we follow \citet{meng2022rome, meng2023memit} and adopt three different metrics: Efficacy Success (ES), which measures the ratio of the number of times the probability of the counter(fact) is higher than that of the original fact i.e. the edit was successful; Paraphrase Success (PS) measures the same but on a paraphrase of the sentence to assess generalisation ability; Neighbourhood Success (NS) tests that semantically similar sentences have not been affected by the edit. 

\begin{table}[h!]
    \centering
    \adjustbox{width = \columnwidth}{
    \begin{tabular}{@{} lccc @{}}
        \toprule
        \textbf{Method} & \textbf{ES $\uparrow$ }& \textbf{PS $\uparrow$} & \textbf{NS $\uparrow$}\\
        \midrule
        LLaMA2 - $D(x)$ & 13.86 & 16.75 & \textbf{83.74} \\
        LoRA - $D_{lora}(x)$ & \textbf{56.89} & 47.05 &  42.83\\
        SFT - $D_{sft}(x)$ & 51.18 & \textbf{51.15} & 48.73 \\

        \midrule
        Ours (fine-tuned) & 56.74 & 50.84 & 43.43 \\
        \bottomrule
    \end{tabular}
    }
    \caption{Results of our method on CounterFact and comparison with baselines.}
    \label{tab:results_counterfact}
\end{table}

Results in Table \ref{tab:results_counterfact} show that our method is competitive with both LoRA and SFT for the CounterFact task, though still presenting a slight degeneration of neighbouring facts (\textbf{NS}) compared to the original PLM. Our results have lower performance than the state-of-the-art methods \cite{meng2023memit, zhang2024comprehensive}, however, they cannot be directly compared. Most methods in knowledge editing outperform fine-tuning baselines, often resorting to much more complex training procedures. In our case, we show how our method is competitive with `less clever' editing methods, confident that their competitiveness would propagate to more complex methods as well.

\subsection{Memory consumption}
\begin{table}[h!]
    \centering
    \adjustbox{max width = \columnwidth}{
        \begin{tabular}{@{} lcccc @{}}
            \toprule
            & \multicolumn{4}{c}{\textbf{Context length}} \\
            \cmidrule{2-5}
            \textbf{Method} & 10 & 512 & 1024 & 2048\\
            \midrule
            $D(\mathcal{K}, x)$ & 13.61 & 16.86 & 20.24 & OOM$^\dagger$ \\
            $E(\mathcal{K}, x)$ & - & 0.52 & 1.24 & 4.1\\
            \midrule
            $D(E(\mathcal{K}, x) + x)$ & - & 14.13 & 14.85 & 17.71 \\
            \bottomrule
        \end{tabular}
    }
    \caption{Peak memory usage in GB when varying on the context length (number of tokens).\footnotesize{$^\dagger$Tested on a NVIDIA RTX 3090 with 24GB of VRAM.}}
    \label{tab:resources}
\end{table}
To further support the computational efficiency of our method, we report in Table~\ref{tab:resources} a comparison of the memory usage of our method compared to using the plain PLM (or LoRA, SFT). The first column with 10 tokens is what we estimate to be the length of an example input, while subsequent columns simulate having knowledge in addition to that input. We show how, while plain PLM runs quickly out of memory (OOM) as the context length increases, our method is still able to run with longer inputs/knowledge. 

Note how, since each encoder is independent from the other, they do not need to be loaded into memory all at once, which further gives memory benefits.

\section{Ablation studies}
Following are our ablation studies. In particular, we experiment with removing a different number of encoders in \ref{sec:less_encoders} and with removing our initial training in \ref{sec:no_pretrain}. For all of our experiments, we train the encoder or fine-tune each method until convergence, all with the same setting.

\subsection{Effects of Having Less Encoders}
\label{sec:less_encoders}
As shown above,  using fewer layers yields the best performance. To corroborate this further, we test our method with three different configurations of encoders. Our training procedure yields the same encoder $E_l$ whether it is trained alone or with any other encoder. Therefore, we can do a single training where we optimise all 32 encoders, one per layer, which allows us to test the performance by removing one or more encoders. We test the following configurations: 
\begin{enumerate}
    \item 32 encoders by having one encoder per layer ($all$).
    \item 28 encoders by only keeping the ones that converged ($conv$).
    \item 6 encoders only by keeping the MEMIT layers ($memit$), the layers that \citet{meng2023memit} found to be the most responsible in storing knowledge: $\{3, 4, 5, 6, 7, 8\}$.
\end{enumerate}

\begin{table}[h!]
    \centering
    \adjustbox{max width = \columnwidth}{
        \begin{tabular}{@{} lcccc @{}}
            \toprule
            & & \multicolumn{2}{c}{\textbf{Time $\downarrow$}} \\
            \cmidrule{3-4}
            \textbf{Layers} & \textbf{Perplexity $\downarrow$} & GPU-hours/epoch & Inference (s)\\
            \midrule
            $memit$ & \textbf{3.33} & \textbf{13.3} & \textbf{268.50}\\
            $conv$ & 3.64 & 50.12 & 629.44\\
            $all$ & 3.74 & 66.2 & 677.39\\
            \bottomrule
        \end{tabular}
    }
    \caption{Comparison of performance, training and inference time when increasing the number of encoders.}
    \label{tab:less_encoders}
\end{table}

We plot in Table \ref{tab:less_encoders} the perplexity of the different combinations of our method on the knowledge augmentation datasets. We can see that keeping only the $memit$ layers outperforms all the other combinations, in both resource consumption and perplexity.

\subsection{Effect of (not) Pretraining the Encoder}
\label{sec:no_pretrain}
It is already possible to notice the effectiveness of our new training objective (Section \ref{sec:training}) from the results in Table \ref{tab:results}, where our method is competitive with the plain PLM while still requiring much fewer resources. However, to further showcase the effectiveness of this pretraining, we `fine-tune' freshly initialised encoders and we compare their performance against our method after pretraining and fine-tuning.
\begin{figure}
    \centering
    \includegraphics[width=\columnwidth]{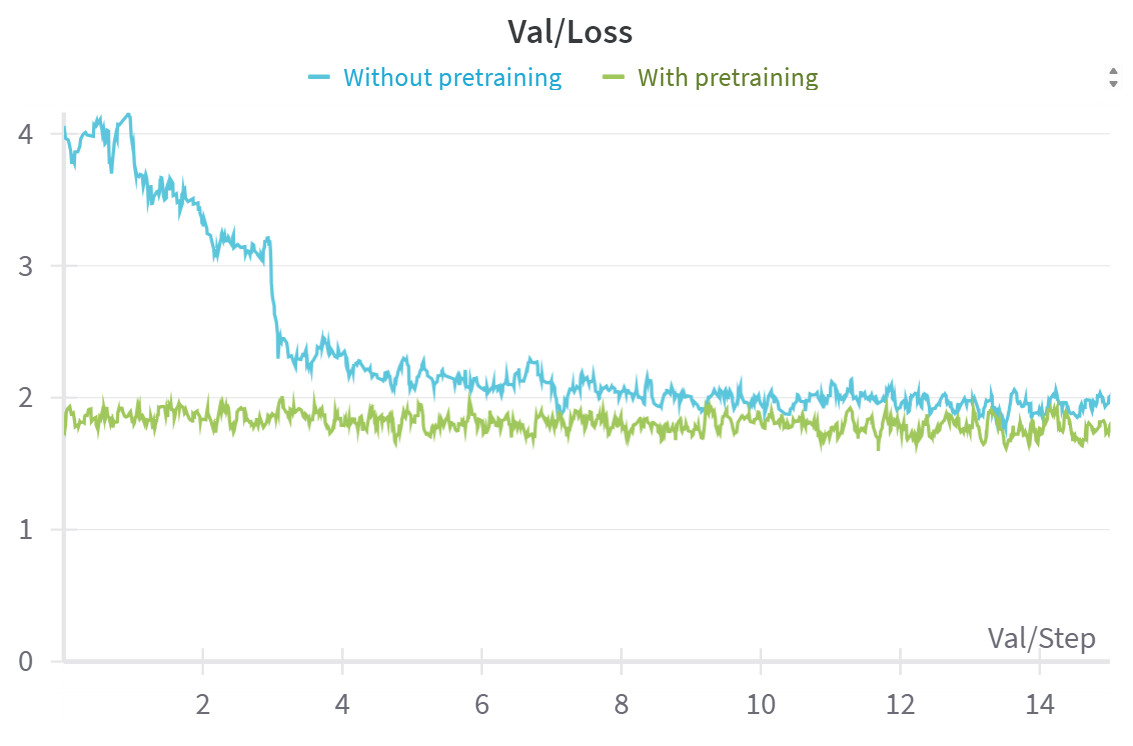}
    \caption{Plot of the validation loss throughout the training of our method with and without pretraining when fine-tuned on the knowledge datasets (all). Note how the validation loss is computed every 250 steps of training.}
    \label{fig:ablation_nopre}
\end{figure}
We report in Figure \ref{fig:ablation_nopre} a comparison of the validation losses. We can already observe how the pretrained model is much quicker in converging, needing only a few examples to adapt to the task, or more specifically $2.89$ epochs compared to $0.84$ for the pretrained version. At the end of the fine-tuning, the pretrained version also outclasses the non-pretrained version which has 3.45 of final perplexity.

\section{Conclusions}
In this work, we have presented DIEK\AE, a method for \textit{efficiently} augmenting and editing knowledge in a PLM. We have also introduced a pretraining objective for DIEK\AE~ that does not rely on back-propagation through the PLM and is, instead, trained by using only its forward passes. We show both the effectiveness of our method and that of our training procedure by comparing it to multiple baselines in knowledge augmentation. Our findings show that our method can yield better results than the plain PLM, while keeping minimal resource consumption ($1/2$ of inference time and $2/3$ of memory). When compared to other baselines, our method is much lighter in training needs (up to $1/7$) but it falls behind in some task performance. Finally, we show how our method can be adapted to the task of knowledge editing, yielding promising results.

% SECTIONS BELOW SHOULD NOT COUNT TOWARDS PAGE LIMIT AND CAN BE UP TO ONE PAGE
\section{Limitations}
Whilst our method brings great benefits in computational costs and performance compared to the plain PLM, it still lags behind in particular tasks, like ICR, where other baselines are much more effective. In terms of knowledge editing, our method is still sub-optimal compared to state-of-the-art methods, thus further research is required to achieve the best results.

Finally, as we stated in Section \ref{sec:encoder}, another limitation of our work is that we do not experiment with any architectural or hyper-parameter changes in the encoders. Our primary contribution does not lie with the model itself but with the integration of the encoders into the PLM and their training procedure.

\section{Ethical Considerations}
Both knowledge editing and augmentation are effective techniques to aid in the generation of PLMs. Such techniques can thus be used for e.g. alleviation of hallucinations, reduction of misinformation and similar. While we did not specifically explore these contexts, our method can be easily adapted to target such problems. Knowledge augmentation, in particular, has also the advantage of allowing better transparency and explainability, where the generations are based on a fixed and given knowledge.

\section*{Acknowledgements}
The computations and data handling were enabled by resources provided by the National Academic Infrastructure for Supercomputing in Sweden (NAISS) at Alvis, C3SE (Chalmers) partially funded by the Swedish Research Council through grant agreement no. 2022-06725.

\bibliography{anthology,custom}

\clearpage
\appendix

\section{Reproducibility details}
\label{app:reproducibility}
This appendix is aimed at explaining and reporting all of the details of our training in order to favour the reproducibility of our results.

\subsection{Half precision}
We adopt mixed precision training full/brain floating point (bf16) for all of our experiments. For inference, all the models are loaded with as bf16.
\subsection{Learning rate}
We start all of our experiments with a learning rate of 1e-5, linearly increasing up to 1e-4, it then decreases back to 1e-5 with a cosine decay.  

For SFT training, this learning rate is too high and results in unstable training so we used 1e-6 instead.
\subsection{Hardware used}
The training jobs were conducted on one or more NVIDIA A100-SXM4-40GB, where necessary (e.g. for fine-tuning or LoRA) we used one or more NVIDIA A100-SXM4-80GB. Note how pretraining our method is the only training that can be done on just a single A100 with 40GB of VRAM, while all the others need at least 80GB of VRAM or double that for SFT. After training, all the evaluations were conducted on a local NVIDIA RTX 3090 (24GB of VRAM). Below are the estimates of the time that each training job took. Each method was trained until convergence unless specificied.
\begin{itemize}
    \item General training of our method when training all layers/encoders (1 epoch): 13h 45m on 2x GPUs
    \item General training of our method when training only 6 layers/encoders (1 epoch): 4h 03m on 2x GPUs
    \item Our method fine-tuned for knowledge augmentation: 2h32m on 2x GPUs
    \item Our method fine-tuned for ICR: 15h 46m on 2x GPU s
    \item Our method fine-tuned for knowledge editing (5 epochs): 2h 20m on 2x GPUs
    \item LoRA baseline for knowledge augmentation: 23h 48m on 2x GPUs
    \item LoRA baseline for ICR: 5h 24m on 2x GPUs
    \item LoRA baseline for knowledge editing (5 epochs): 1h 14m on 2x GPUs
    \item SFT baseline for knowledge augmentation: 31h 22m on 2x GPUs
    \item SFT baseline for ICR: 14h 40m on 2x GPUs
    \item SFT baseline for knowledge editing (5 epochs): 1h 45m on 2x GPUs
\end{itemize}

\subsection{LoRA baseline}
As written in the main paper, for our LoRA baseline we adopt a comparable number of trainable parameters to the method we're comparing to. In particular, when comparing to our fine-tuned method on only MEMIT layers (layers 3,4,5,6,7,8) we use a rank $r=24$ for LoRA. Other hyper-parameters we used are $\alpha=32$ and 0.05 for the dropout.

\section{Pretraining stability}
\label{app:enc_cov}
\begin{figure*}
    \begin{subfigure}{.5\textwidth}
          \centering
          \includegraphics[width=\columnwidth]{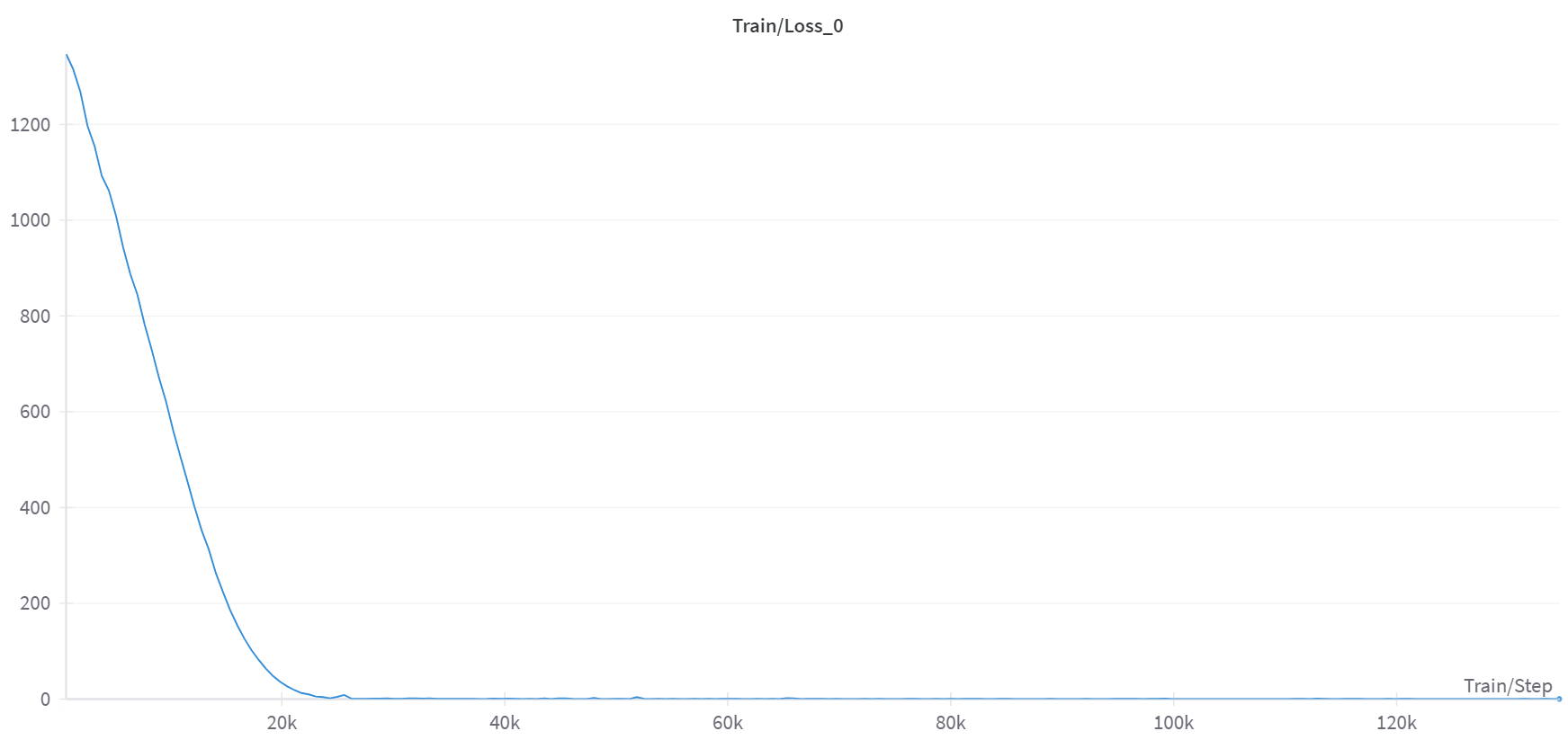}
          \label{fig:loss_1}
    \end{subfigure}%
    \begin{subfigure}{.5\textwidth}
          \centering
          \includegraphics[width=\columnwidth]{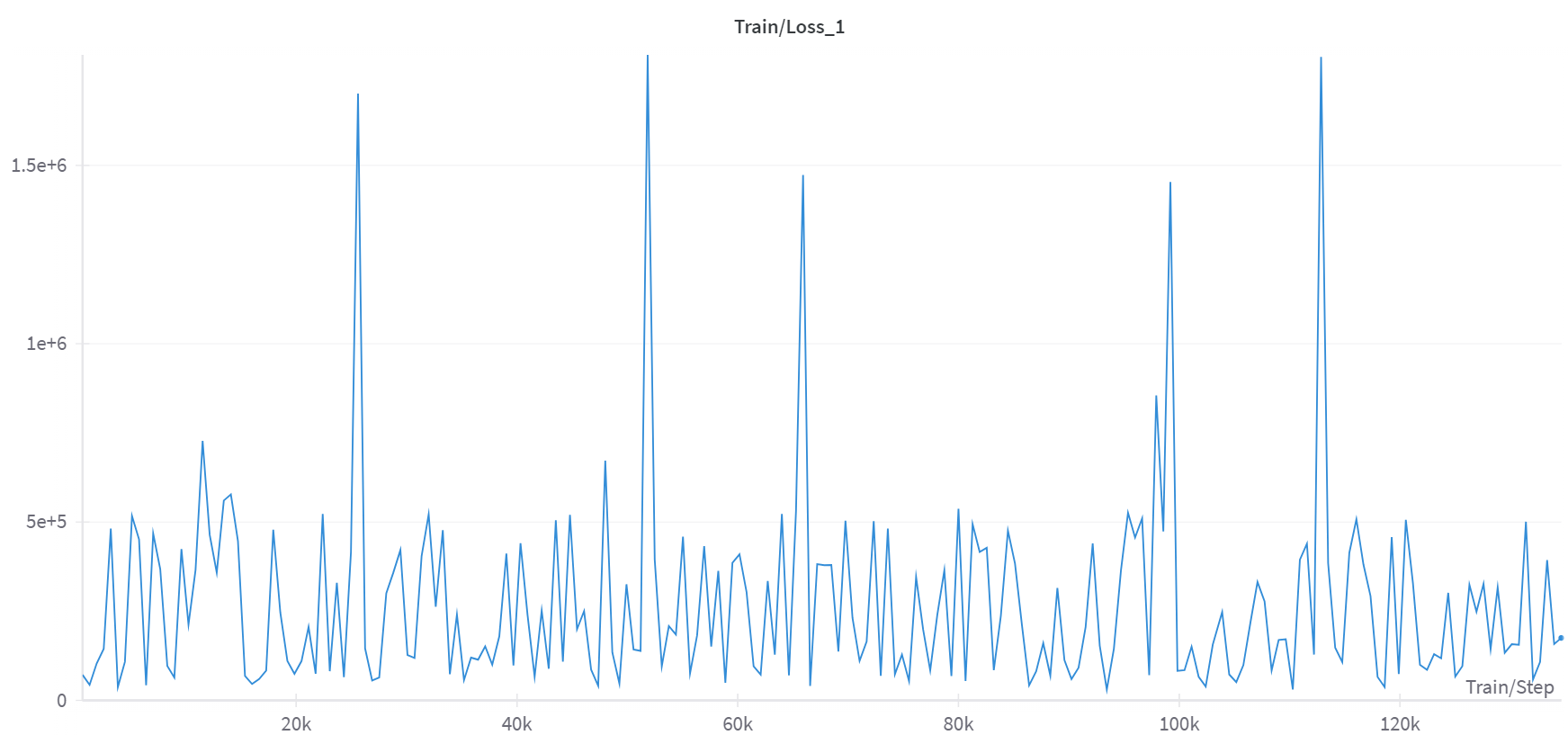}
          \label{fig:sub2}
    \end{subfigure}
    \caption{Pretraining loss of a layer that converges (left) and one that does not (right).}
    \label{fig:converg}
\end{figure*}

As anticipated in the main paper, not all of the encoders are able to converge and learn from the training objective that we set. In particular, we find that those relative to the decoder's layer $\{1, 29, 30, 31\}$ do not improve at all during training (also see Figure \ref{fig:converg}).

While it is understandable that the last three layers are harder to learn due to the very high variability involved; we cannot make any educated guess about the second layer, which is truly an outlier as the first layer is correctly learned.

We hypothesise this also being partially due to the architecture and partially to the training. In previous tests on the first generation of LLaMA the layers that did not converge were $\{2, 30, 31\}$.

\section{Datasets preprocessing}
In this section, we provide more information on the datasets that we used and how we preprocessed them. When possible, we use the train/val/test split from the original dataset. Table \ref{tab:datasets} shows some examples from each dataset. When we process the dataset, we include in the input the tokens `[USR]' for a sentence that is relative to the user and `[SYS]' for sentences from the system (language model). When computing the losses we only consider the sentences from `[SYS]'. Further, when more than one `knowledge' is provided for a single input-output, we include all the knowledge separated by another token `[SEP]'.

\label{app:data_preprocessing}
\begin{table*}[h!]
    \centering
    \begin{tabular}{|p{0.11\textwidth}||p{0.36\textwidth}|p{0.26\textwidth}|p{0.18\textwidth}|}
        \hline
        Dataset & Knowledge & Source & Target\\
        \hline
        \hline
        CMU Dog   & Genius, billionaire, and playboy Tony Stark, who has inherited the defense contractor Stark Industries [...] & In your opinion was Robert Downey Jr a great fit to play this role? & His performance of a wild and egotistical playboy was very [...]\\
        \hline
        Curiosity & The establishment of the ``State of Palestine'' was announced by the PLO legislature in November 1988 [...] & What about security in Palestine? & [...] In 1988, the PLO [...]\\
        \hline
        Dream & Woman: Can I help you? Man: I'm looking for some suit that I can [...] & Where does the conversation most probably take place? & In a clothing store.\\
        \hline
        Natural Questions & The Shannara Chronicles Genre Fantasy Created by Alfred Gough Miles [...] & What is the shannara chronicles season 2 based on? & The Sword of Shannara Trilogy \\
        \hline
        QUASAR-T & However, the largest city in Scotland in terms of population is Glasgow. Like Chicago is not [...] & Which city is, in terms of population, the second largest in Mexico? &  Guadalajara\\
        \hline
        Wizard of Wikipedia & My favorite artist is guns and roses. Guns N' Roses, often abbreviated as GNR, is an American hard rock [...] & I just recently started listening to Guns N' Roses. & Oh they are classic!! They are an American band [...]\\
        \hline
        \hline
        ProofWriter & he bear eats the dog. The bear eats the squirrel. The bear is green. The bear is red. The bear sees the rabbit [...] & The bear sees the rabbit. & True\\
        \hline
        Counterfact & Nokia (original fact) & Nokia N900 is produced by & IBM (new counter-fact) \\
        \hline
    \end{tabular}
    \caption{Example knowledge, input, output from each dataset.}
    \label{tab:datasets}
\end{table*}

CMU Dog \cite{zhou-etal-2018-dataset}, Curiosity \cite{rodriguez-etal-2020-information}, DREAM \cite{sun-etal-2019-dream}, Natural Questions \cite{kwiatkowski-etal-2019-natural}, QUASAR-T \cite{dhingra2017quasar}, Wizard of Wikipedia \cite{dinan2018wizard}.
\subsection{CMU dog}
CMU Dog \cite{zhou-etal-2018-dataset} contains conversations between two users based on a given snippet of film information. The conversation generally begins by talking about the film in general e.g. cast, name, etc. and then continues with plot, etc. Each conversation is also rated based on its quality on a scale of up to 3. When we preprocess the dataset we exclude any sample with a rating lower than 3. The knowledge is given by the information of the film, input/output are the single messages of the two users.

After preprocessing we have 15665/1515/2403 for the train/val/test split.

\subsection{Curiosity}
Curiosity \cite{rodriguez-etal-2020-information} contains conversations between an `assistant' and a user. The assistant goal is to reply to the user's query by including relevant `fun' facts (the knowledge). Note how, sometimes, the assistant would not use \textit{any} facts, in such cases, we take as knowledge the assistant's reply. In order to avoid having the knowledge the same as the target, in such cases, we paraphrase the knowledge. 

After preprocessing we have 65624/8588/8584 for the train/val/test split.

\subsection{DREAM}
DREAM \cite{sun-etal-2019-dream} is a dataset aimed at reading comprehension. Each sample contains a conversation between two people, alongside a question on that text.

After preprocessing we have 3829/1273/1276 for the train/val/test split.

\subsection{Natural Questions}
Natural Questions \cite{kwiatkowski-etal-2019-natural} is a dataset on open-domain question answering with a given knowledge for each question-answer pair. Given the size of the original dataset (42GB), we rely on a preprocessed version from HuggingFace\footnote{\url{https://huggingface.co/datasets/cjlovering/natural-questions-short}}, which however, does not contain any test set. For this reason, we do not include this dataset in the evaluation results in the main paper's Table \ref{tab:results}

After preprocessing we have 13933/871/0 for the train/val/test split.
\subsection{QUASAR-T}
% TODO: also include number of samples in the dataset
QUASAR-T \cite{dhingra2017quasar} is another dataset aimed at question answering, which also includes the results from a web search as knowledge.

Note how this dataset's knowledge does not always contain facts relevant to the question, as we showcase in Table \ref{tab:datasets}.

After preprocessing we have 37012/3000/3000 for the train/val/test split.

\subsection{Wizard of Wikipedia}
Wizard of Wikipedia \cite{dinan2018wizard} is a dataset containing the interactions between a user and a `wizard'. The wizard's goal is to reply to the queries of the user with the given knowledge.

After preprocessing we have 70438/3779/3544 for the train/val/test split.

\subsection{Counterfact}
Counterfact is a dataset created by \citet{meng2022rome} by swapping objects of sentences. As an example, consider the sentence ``The mother tongue of Danielle Darrieux is French'', with the object `French', this can be swapped with a different language e.g. `English' to obtain a counterfactual sentence.

After preprocessing we have 18631/3288/0 for the train/val/test split.

\subsection{ProofWriter}
ProofWriter \cite{tafjord-etal-2021-proofwriter} is a dataset about logical reasoning. The knowledge is composed of facts and rules (that are not universally true, e.g. ``the dog is white'') and a series of assertions that can be True/False/Unknown given the knowledge. We only use the `depth-2' part of the dataset, meaning that each statement can be classified with 2 `hops'. In the knowledge we also keep all the irrelevant facts for the statement.

After preprocessing we have 70076/10094/19840 for the train/val/test split.

\subsection{FOLIO}
FOLIO \cite{han2022folio} is a dataset about first order logical reasoning. The knowledge is composed of facts and rules and a series of assertions that can be True/False/Unknown given the knowledge. 

After preprocessing we have 1004/204/0 for the train/val/test split.
\end{document}